\documentclass{article} 
\usepackage[utf8]{inputenc} 
\usepackage{amsmath,amssymb,amsfonts} 
\usepackage{graphicx} 
\usepackage{authblk} 
\usepackage[left=2cm, right=2cm, top=2cm]{geometry}
\usepackage{authblk}

\title{Generative Pretraining at Scale: Transformer-Based Encoding of Transactional Behavior for Fraud Detection }

\author[1,*]{Ze Yu Zhao}
\author[1,*]{Zheng Zhu}
\author[1,$\dagger$]{Guilin Li}
\author[1,$\dagger$]{Wenhan Wang}
\author[1]{Bo Wang}
\affil[ ]{{feynmanzhao, adamzzhu, guilinli, ezewang, pollowang}@tencent.com}
\affil[1]{Tencent, WeChat Pay}
\date{}
\begin{document}

\maketitle
\renewcommand{\thefootnote}{}%
\footnotetext{* Co-first authors with equal contributions $\dagger$ Corresponding authors}
\renewcommand{\thefootnote}{\arabic{footnote}}%
\setcounter{footnote}{0}%
 
\begin{abstract}

In this work, we introduce an innovative autoregressive model leveraging Generative Pretrained Transformer (GPT) architectures, tailored for fraud detection in payment systems. Our approach innovatively confronts token explosion and reconstructs behavioral sequences, providing a nuanced understanding of transactional behavior through temporal and contextual analysis. Utilizing unsupervised pretraining, our model excels in feature representation without the need for labeled data. Additionally, we integrate a differential convolutional approach to enhance anomaly detection, bolstering the security and efficacy of one of the largest online payment merchants in China. The scalability and adaptability of our model promise broad applicability in various transactional contexts.
\end{abstract}

\section{Introduction}

In the dynamic world of digital finance, protecting transactions from fraud is an increasingly intricate challenge. Machine learning has advanced behavioral analysis, yet it faces difficulties with the detailed patterns of transaction data—both vast in volume and complex in nature. Conventional models fall short in decoding this data accurately.

Our study introduces a novel method that applies Generative Pretrained Transformers, acclaimed for language understanding, to model financial transactions. By pretraining on extensive user payment data, our model overcomes the common obstacles of behavioral sequence analysis such as the need for large amounts of labeled data.

We detail three major contributions: firstly, an innovative autoregressive pretraining method designed specifically for financial data. Secondly, we introduce a differential convolution structure that enhances the model's capacity for anomaly detection. Lastly, from a methodological perspective, we demonstrate the model's flexibility in adapting to diverse financial scenarios, and from an application standpoint, we validate its scalability through extensive testing in our dataset.

Our work not only moves the needle forward in financial fraud detection but also showcases the potential of unsupervised learning in environments where protecting data and user privacy is crucial.

\section{Related Work}
\subsection{Sequential Recommendation and Pre-training Techniques}
Sequential recommendation models primarily rely on users' chronological behavior sequences to learn their preferences. Recently, various deep neural networks have been employed for sequence-based recommendation, such as GRU4Rec \cite{1511.06939}, which uses Gated Recurrent Units for session-based recommendation. Inspired by the success of Transformer \cite{1706.03762} and BERT \cite{1810.04805}, models like SASRec \cite{1808.09781} and Bert4Rec \cite{1904.06690} adopt self-attention mechanisms to model user behavior sequences. Cold-start problems in sequential recommendation have also attracted researchers' attention, with methods like meta-learning mechanisms and behavior sequence augmentation being proposed to address these issues.

Pre-training techniques have achieved significant success in NLP and CV, aiming to learn prior knowledge from large-scale datasets to aid specific downstream tasks. After pre-training, models are further fine-tuned on downstream supervised signals to fit the specific task. With the thriving of pre-training, many pre-training models have been proposed in recommendation, such as BERT4Rec \cite{1904.06690}, S3Rec \cite{2008.07873}, \cite{2010.14395}, PeterRec \cite{2001.04253}, and \cite{2102.10989}. These models employ various pre-training strategies, such as masked item prediction, contrastive learning, and incorporating user profiles and social relations.

Some recent works have attempted to introduce pre-trained language models (PLMs) into recommendation systems, such as UniSRec\cite{2206.05941}, P5\cite{2203.13366}, and M6-Rec\cite{2205.08084}. These models leverage PLMs to learn universal item representations across different domains using items' textual information or generate textual explanations for recommendations. However, these models often rely on additional textual information and PLMs, making them less universal and flexible.

In light of these developments, we propose a novel approach that aims to (a) bridge the gap between pre-training and downstream recommendation tasks, and (b) better extract useful personalized knowledge from pre-trained models by replacing fine-tuning with personalized prompt-tuning. Our approach is more universal and flexible, as it does not depend on additional textual information and PLMs. We also demonstrate the effectiveness of our method across different tasks and pre-training models.

\subsection{Relationship between Data Compression and Behavioral Sequence Models}

Data compression is not only relevant to pre-trained models in general but also has a direct connection to behavioral sequence models. Behavioral sequence models aim to learn and predict user behavior patterns based on their historical interactions. By leveraging data compression, these models can effectively capture the essential features and temporal dependencies in user behavior data, leading to improved performance in various applications, such as recommendation systems and risk control.

The relationship between data compression and behavioral sequence models can be explained through the following aspects:

1. \textit{Compact Representation}: Data compression encourages the development of compact and efficient representations of user behavior sequences. By compressing the input data, behavioral sequence models can capture the most relevant and essential features, making it easier to learn the underlying patterns and dependencies in user behavior data.

2. \textit{Noise Reduction}: Behavioral sequence data often contains noise and irrelevant information that can hinder the learning process. Data compression helps behavioral sequence models filter out this noise and focus on the most important features, leading to more accurate and robust predictions.

3. \textit{Temporal Dependency Learning}: One of the key challenges in behavioral sequence modeling is capturing the temporal dependencies between user actions. Data compression encourages the development of models that can efficiently learn and represent these dependencies, improving the model's ability to predict future user behavior based on past interactions.

4. \textit{Scalability}: Data compression promotes the development of models that can handle large-scale datasets and complex behavioral patterns. By compressing the input data and learning efficient representations, behavioral sequence models can scale to handle large numbers of users and items, making them more suitable for real-world applications.

In summary, the relationship between data compression and behavioral sequence models is crucial for improving the performance of these models in various applications. By leveraging the principles of data compression, behavioral sequence models can learn compact and efficient representations of user behavior data, capture temporal dependencies, reduce noise, and scale to handle large-scale datasets. This improved performance can ultimately lead to better recommendations, more accurate risk control, and enhanced user experiences in a wide range of applications.

\section{Pretrain Data Selection and Processing}
For the pretraining phase, it is crucial to select adequate user behavioral data to avoid data sparsity and facilitate the learning process. We utilized our user data, one of China's largest payment platforms, and selected the data based on the following criteria:

1. Focus on users with high transaction frequency, as they exhibit more consistent and characteristic behavior patterns.

2. Include all user sequences reported with abnormalities.

3. Randomly extract a specific time window from users' sequences and sample within a longer time window.

These criteria help us obtain a realistic representation of payment patterns. The real-world payment sequences enhance the pretraining model's capacity to learn and generalize user behavior patterns, improving its predictive capabilities.

Our dataset comprises payment records of roughly 200 million users, totaling around 15 billion payment behavior tokens. We consider some distinct features, resulting in over 1.3 trillion tokens. These features, such as payment amount, time, and channel, have been converted into multivariate time-series data, making them suitable for analysis and learning by sequential models.

\section{Autoregressive Structure for Sequential Behavior Modeling}

Behavior sequence models are inherently temporal, making autoregressive structures, like the GPT\cite{2006.15720} model, more suitable than bidirectional attention mechanisms used in BERT. Autoregressive structures align with the temporal nature of user behavior sequences, capturing dependencies between past and future actions while preserving event order.

The probability of observing a user behavior sequence, $B_1, B_2, \dots, B_T$, in a GPT-based generative model can be factorized as the product of conditional probabilities:

$$
P(B_1, B_2, \dots, B_T) = \prod_{t=1}^{T} P(B_t | B_{t-1}, B_{t-2}, \dots, B_{1})
$$

This factorization aligns with the autoregressive structure, capturing the influence of past behaviors on the current behavior.

Comparing the GPT-based generative model formula with the autoregressive (AR) time series model formula:

$$
X_t = c + \sum_{i=1}^{p} \phi_i X_{t-i} + \epsilon_t
$$

both models share the core idea of modeling the current observation (or user behavior) based on previous observations. The AR time series model uses a linear combination of lagged observations with coefficients $\phi_i$, while the GPT-based generative model computes conditional probabilities of the current observation given past observations.

These similarities demonstrate the applicability of the GPT structure to behavior sequence modeling, effectively capturing temporal dependencies and contextual relationships in user behavior data. The autoregressive structure is more suitable for modeling and predicting user behaviors sequentially compared to bidirectional attention mechanisms.

\section{A Comprehensive Pretraining Framework for Sequential Behavior Modeling}
\begin{figure}[ht]
\centering
\includegraphics[width=0.7\textwidth]{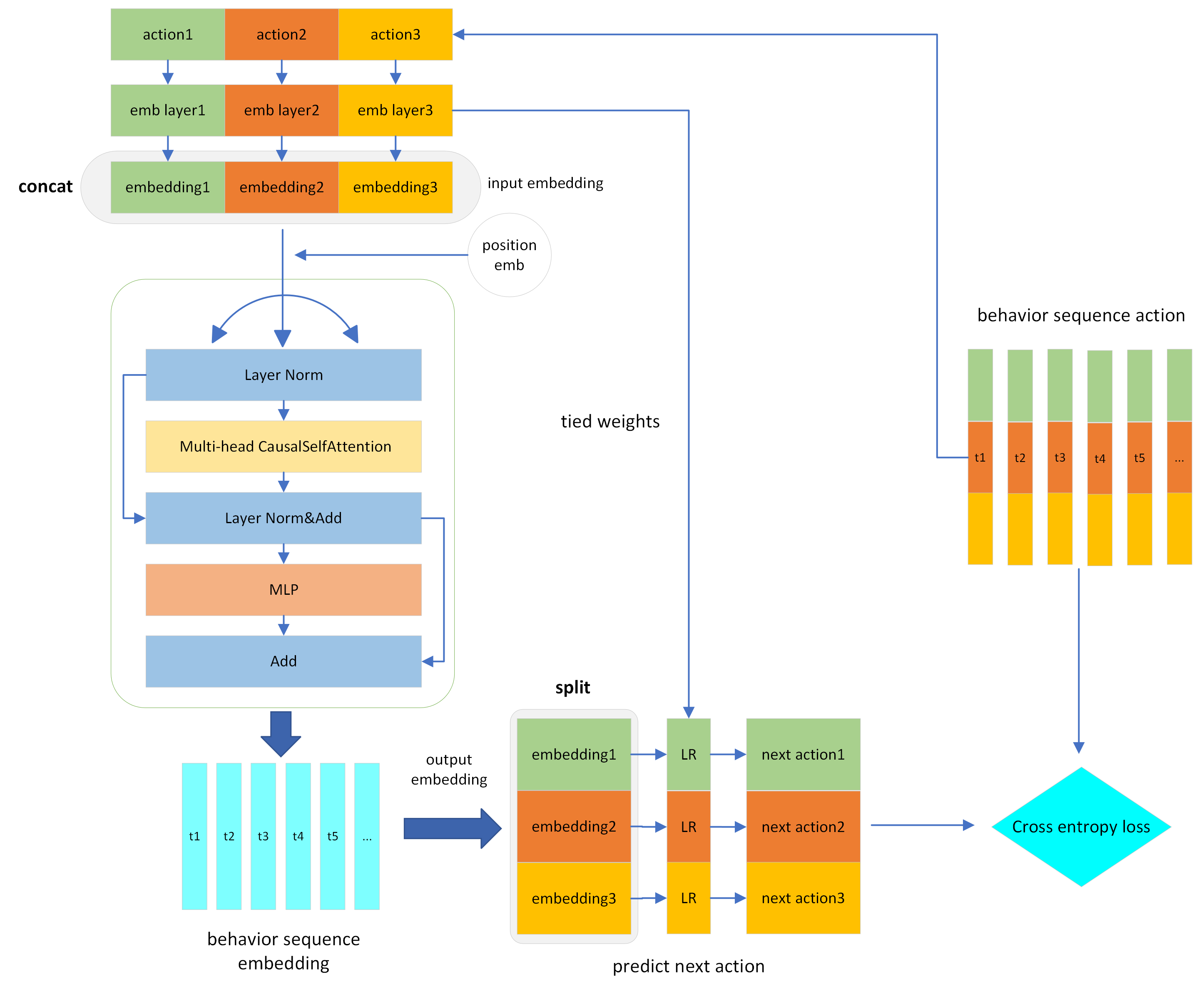}
\caption{Pre-training models.}
\label{fig:pretraining}
\end{figure}

In this section, we propose a novel pretraining framework for sequential behavior modeling. The framework is designed to effectively capture the temporal dependencies and contextual relationships in user behavior data while addressing the challenges of token explosion and behavior reconstruction.

\subsection{Multivariate Sequence Embedding}

Given a user behavior sequence at time point $t$, the behavior $a_t$ is characterized by multiple discrete attributes $a_t = [a_1, a_2, a_3, \dots]_t$. We employ an embedding layer to transform these multivariate behavior sequences into a continuous representation. These embeddings are then concatenated chronologically, forming an input matrix of shape (sequence length, dim).

\subsection{Autoregressive Modeling of Sequential Behaviors}

The proposed framework leverages an autoregressive structure to model temporal dependencies in user behavior sequences. The objective is to minimize the cross-entropy loss for predicting the next behavior, given the previous behaviors. Formally, the model learns the probability distribution of the $t$-th behavior, $P(a_t | a_{t-n}, \dots, a_{t-1}; \theta)$, conditioned on the previous $n$ behaviors in the sequence.

\subsection{Addressing Token Explosion with Concatenation}

To mitigate token explosion in multivariate behavior sequences, we introduce a concatenation operation for token embeddings. This approach allows different dimensions of behavior tokens to be combined, forming a compact representation of the behavior state at time $t$. Consequently, the embedding layer's dimensionality is reduced, alleviating sparsity and enhancing generalization capabilities.

\subsection{Behavior Reconstruction Loss for Predict-Next Task}

To predict the next behavior in the sequence, we propose a behavior sequence reconstruction strategy. Instead of predicting a single token, the model is trained to reconstruct the next behavior by splitting the output embedding into multiple corresponding representation spaces. Each representation space expresses a particular aspect of the next behavior. This reconstruction task enables the model to predict possible future states, facilitating autoregressive learning.

Let the cross-entropy loss of each behavior token be averaged so that each dimension can be learned simultaneously:

$$
\mathcal{L}_{\text{reconstruction}} = \frac{1}{T}\sum_{t=1}^{T} \frac{1}{D}\sum_{d=1}^{D} \mathcal{L}_{\text{CE}}(a_{t,d}, \hat{a}_{t,d})
$$

where $T$ is the sequence length, $D$ is the number of dimensions, $a_{t,d}$ is the true token at time $t$ and dimension $d$, and $\hat{a}_{t,d}$ is the predicted token.

\subsection{Weight Tying for Parameter Sharing}

Our framework employs weight tying to implement parameter sharing between the input and output embedding layers in the autoregressive model. This technique offers two main advantages:

1. It enhances training efficiency by reducing the number of learnable parameters.
2. It serves as a regularization method, stabilizing the weight updates of the embedding layer and ensuring symmetrical unification between the encoding and decoding processes.

In summary, the proposed pretraining framework for sequential behavior modeling addresses the challenges of token explosion and behavior reconstruction while effectively capturing the temporal dependencies and contextual relationships in user behavior data. The autoregressive structure, combined with the novel techniques introduced in this framework, makes it a promising approach for modeling and predicting user behaviors in a sequential manner.

\section{Fine-Tuning for Anomaly Detection with Differential and Convolutional Operations}
\begin{figure}[ht]
\centering
\includegraphics[width=0.7\textwidth]{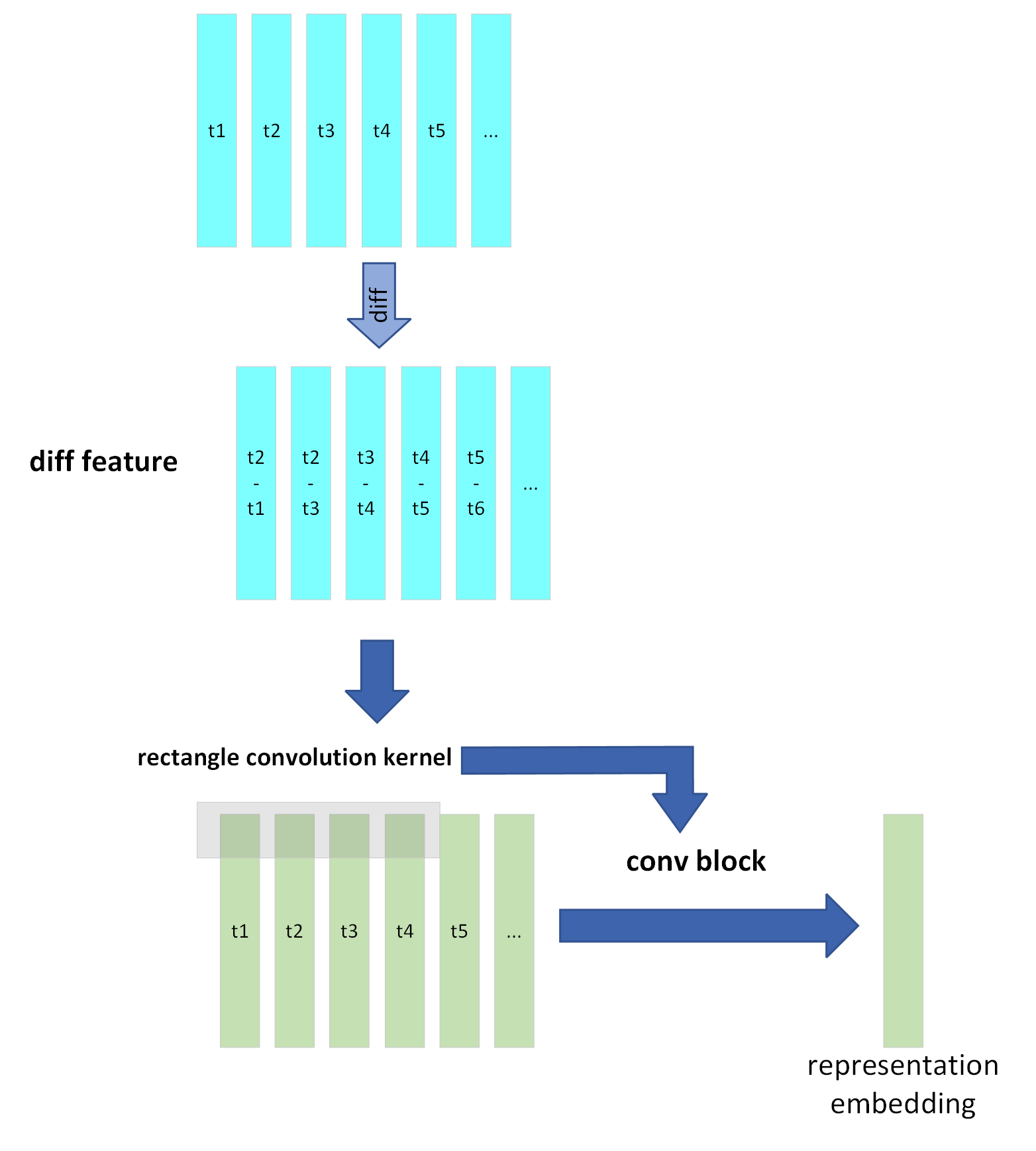}
\caption{Different encoder finetune.}
\label{fig:pretraining}
\end{figure}

After pretraining, the model needs to be fine-tuned for downstream tasks, such as anomaly detection. To achieve this, the model should capture contextual differences within each sequence and compress the 2D sequence into a 1D representation for detecting anomalous changes.

\subsection{Supervised Fine-Tuning (SFT)}

\subsubsection{Differential Operation for Temporal Stationarity}

We introduce a sequence tensor differential method to capture fluctuations, inspired by the differential operation in time series domain. This method eliminates non-stationarity, reduces the impact of long-term trends and seasonality, and is computationally efficient. We use the first-order difference to optimize the prediction sequence and highlight the influence of anomaly points on the behavior embedding.

\subsubsection{Convolutional Operation for Anomaly Detection}

After obtaining the difference sequence, we employ a convolutional approach to construct an anomaly perception layer. This approach offers local feature extraction, parameter sharing, translation invariance, multi-scale feature extraction, and automatic feature learning. These advantages enable the model to effectively detect anomalous patterns in the sequence data, enhancing the flexibility and robustness of anomaly detection.

In summary, we use a multi-layer convolutional approach to represent sequence anomalies, serving as the core method for fine-tuning the model on downstream tasks.

\section{Unsupervised Fine-Tuning via Contrastive Learning}

\begin{figure}[ht]
\centering
\includegraphics[width=0.5\textwidth]{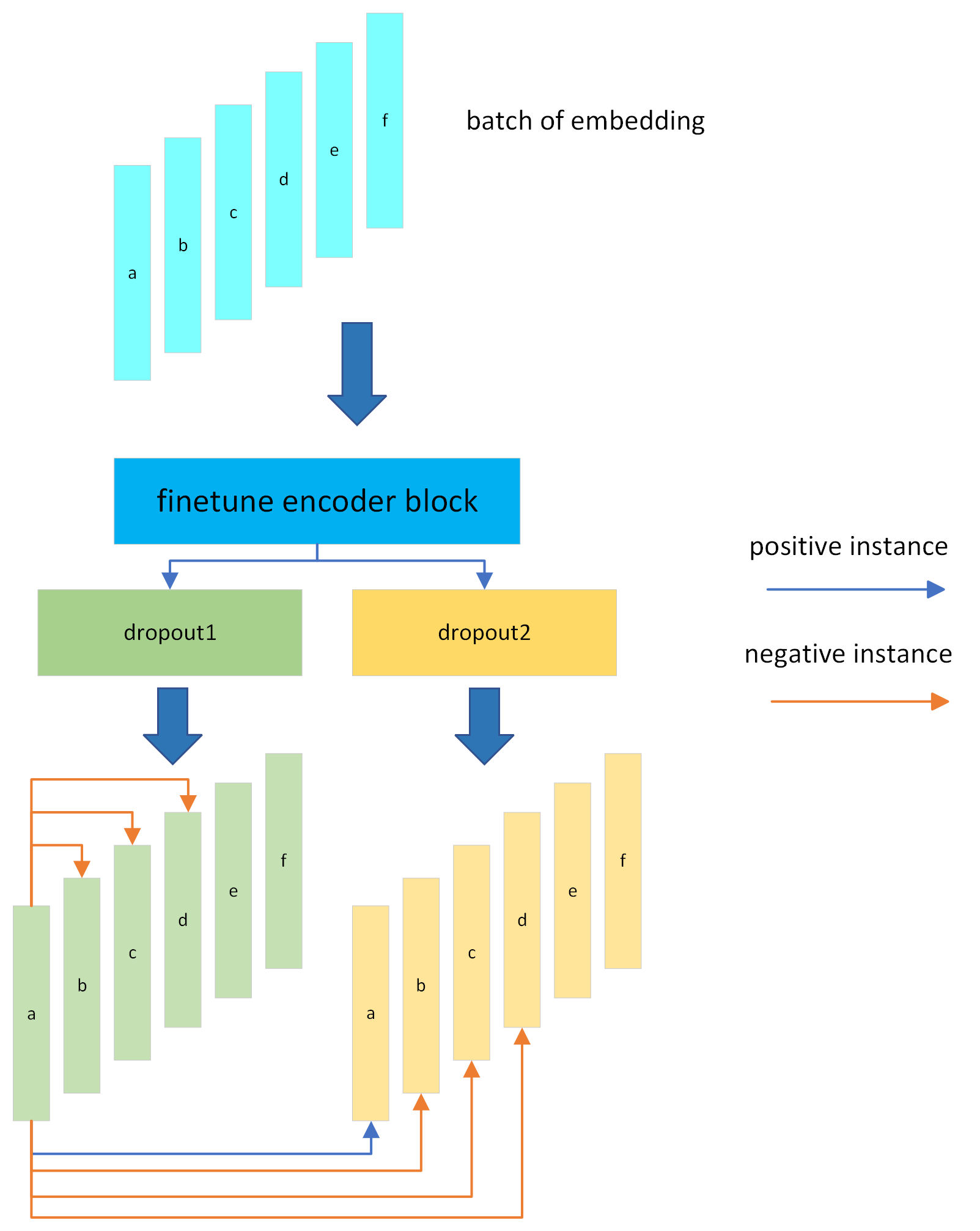}
\caption{ Contrastive learning}
\label{fig: contrastive learning}
\end{figure}

\subsection{Contrastive Learning}

Contrastive learning is a powerful unsupervised learning technique that achieves success in domains like natural language processing (NLP) and computer vision (CV). By incorporating contrastive learning into pretrained models, we can exploit unlabeled data for fine-tuning, improving performance on downstream tasks.

Contrastive learning focuses on learning meaningful representations that enable the model to distinguish between similar and dissimilar data points. This is achieved by optimizing a contrastive loss function, which encourages the model to minimize the distance between positive pairs (similar instances) and maximize the distance between negative pairs (dissimilar instances) in the learned feature space.

We adopt a dropout-based method, akin to SimCSE, to construct contrastive negative samples. This method learns to discern differences between sequence patterns, demonstrating strong generalization capabilities. The core idea is to use dropout as a data augmentation technique, generating multiple views of the same instance, which serve as positive pairs for contrastive learning.

\subsection{Contrastive Loss for Behavior Sequence Embeddings}

To perform contrastive learning on behavior sequence embeddings, we need a suitable loss function that encourages the model to learn meaningful representations. A commonly used loss function is the InfoNCE loss, defined as:

$$
\mathcal{L}_i = - \log \frac{\exp\left(\text{sim}(v_i, v_i^+)\right)}{\exp\left(\text{sim}(v_i, v_i^+)\right) + \sum_{j=1}^{N-1} \exp\left(\text{sim}(v_i, v_i^j)\right)}
$$

where $\text{sim}(v_i, v_j)$ is a similarity function that measures the similarity between embeddings $v_i$ and $v_j$, such as the cosine similarity.

The overall InfoNCE loss for all behavior sequence embeddings is given by:

$$
\mathcal{L} = \frac{1}{N} \sum_{i=1}^{N} \mathcal{L}_i
$$

The model is trained to minimize this loss, learning embeddings that are similar for positive pairs and dissimilar for negative pairs. By optimizing the contrastive loss, the model captures the intrinsic structure and patterns in the behavior sequence data, leading to more meaningful and discriminative embeddings.

\section{Exploring Few-Shot Learning in Anomaly Detection}
\begin{figure}[ht]
\centering
\includegraphics[width=0.5\textwidth]{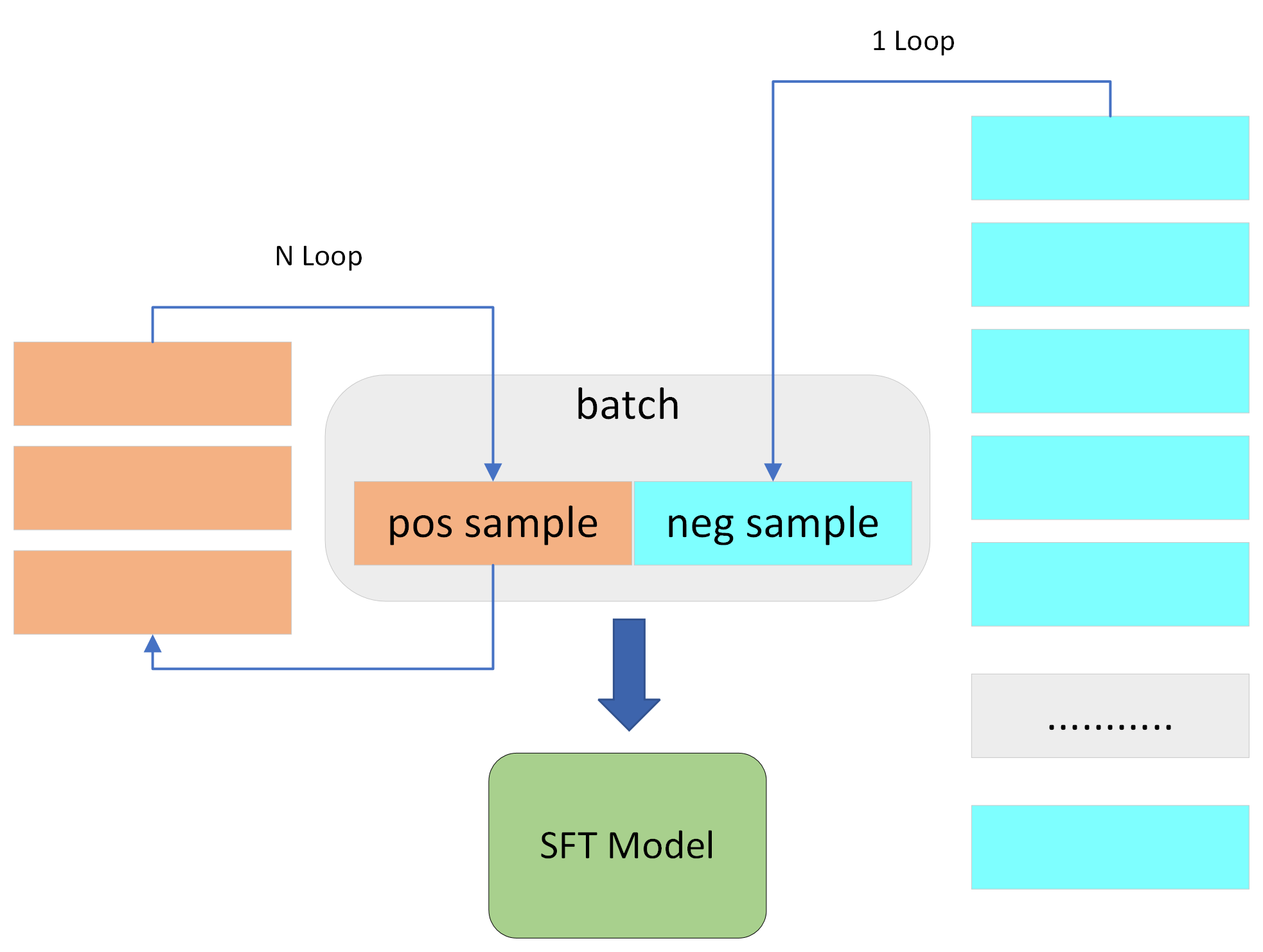}
\caption{ Contrastive learning}
\label{fig: contrastive learning}
\end{figure}

We investigate the potential of few-shot learning\cite{2005.14165} for anomaly detection in payment risk control systems. By leveraging the generalization capabilities of pretrained models, we address the challenges associated with limited positive samples.

\subsection{Relationship between Anomaly Detection and Few-Shot Learning}

Payment risk control systems rely on user complaints and reports to identify anomalous accounts. However, reported accounts represent only a small fraction of the total anomalous accounts, making the risk control system a "semi-supervised" learning task. The model's generalization ability is crucial for its performance.

Pretrained models hold great potential for few-shot learning due to their exposure to vast amounts of data. Few-shot learning focuses on the model's ability to "learn from few examples" and generalize to unseen tasks. The more patterns and information the model memorizes during pretraining, the more adaptable it becomes to a diverse range of downstream tasks.

\subsection{Importance of Anomaly Detection in Payment Risk Control Systems}

A stable payment risk control system relies on detecting and preventing potential anomalous behaviors in advance. It is crucial for the model to learn latent anomalous behaviors from limited available data on detected anomalies.

We deploy the model in an offline manner, focusing on monitoring anomalous account behaviors during normal operation, enabling early detection and prevention of malicious accounts.

\subsection{Few-Shot Supervised Fine-Tuning with Imbalanced Sampling}

Due to the scarcity of positive samples compared to the abundance of negative samples, we adopt different sampling strategies for positive and negative samples during supervised fine-tuning.

For limited positive samples, we employ repetitive sampling, while for abundant negative samples, we use random full sampling. This allows the model to thoroughly compare differences between limited positive samples and massive negative samples during the fine-tuning process, enhancing its ability to distinguish between anomalous and non-anomalous behaviors.

In conclusion, leveraging pretrained models' generalization capabilities and employing few-shot learning techniques address the challenges associated with the scarcity of positive samples in payment risk control systems. The proposed approach enables the model to learn latent anomalous behaviors from limited available data and generalize to a diverse range of downstream tasks, contributing to the payment risk control system's stability and effectiveness.

\section{Experiments}

In this section, we present experimental results for various tasks, including multi-class classification, binary classification scoring, and contrastive learning-based embedding retrieval. we only used 9 behavior features for the experiments.

\subsection{Multi-Class Classification Results}

We evaluate the model's performance using a multi-class classification task. Fraudulent behaviors in the payment ecosystem are extremely sparse samples, making them well-suited for few-shot learning scenarios. Many fraud categories have only a few hundred samples, which is a minuscule fraction of the massive user base.

Our experimental data includes a validation set composed of 500,000 negative samples and 22,042 positive samples. Importantly, these positive samples have not been subjected to rigorous manual review but were compiled from complaint records, which may result in the inclusion of false positives. Multiclass tasks require more positive sample data, and manually screening this data can be costly. We plan to provide more manually screened data in subsequent experiments.

The test results are as follows:

\begin{table}[h]
\centering
\begin{tabular}{|l|l|l|l|}
\hline
Label & Recall (\%) & Precision (\%) & Eval Positive Ratio (\%) \\ \hline
Normal & 98.7 & 99.0 & 96.1 \\ \hline
Free Gift Fraud & 13.6 & 30.6 & 0.1 \\ \hline
Adult Content Fraud & 1.3 & 23.6 & 0.18 \\ \hline
Undelivered Transaction & 19.6 & 27.9 & 0.58 \\ \hline
Gaming Fraud & 4.2 & 25.8 & 0.35 \\ \hline
Other Fraud & 28.5 & 34.6 & 0.27 \\ \hline
Financial Credit Fraud & 5.14 & 27.2 & 0.077 \\ \hline
Part-Time Job Fraud & 59.6 & 49.5 & 0.34 \\ \hline
Dating Fraud & 76.2 & 42.0 & 1.34 \\ \hline
\end{tabular}
\caption{Multi-class classification results}
\end{table}

We observe that although "Part-Time Job Fraud" does not have the highest number of samples, it achieves the highest precision and second-highest recall. This indicates that the model has a strong ability to detect anomalous patterns that emerge from sudden changes in normal account behaviors.

\subsection{Abnormal Behavior Scoring for Binary Classification}

Next, we test the model's performance in assigning abnormal behavior scores to users in a binary classification task.

Test methodology: Among 80 million large-transaction users, there are over 800 anomalous accounts. We assign abnormal behavior scores to these users and test the model's effectiveness by ranking the scores in descending order. Positive samples account for 0.001\% of the dataset.

Please note that the positive samples in this case are carefully selected through manual screening of complaint samples. These complaints are absolutely genuine fraud cases, and there are no false positives.

\begin{table}[h]
\centering
\begin{tabular}{|l|l|l|}
\hline
Rank & Precision (\%) & Recall (\%) \\ \hline
Top 1\% & 0.02& 19.16 \\ \hline
Top 0.1\% & 0.05& 5.5 \\ \hline
Top 0.01\% & 0.13& 1.19 \\ \hline
\end{tabular}
\caption{Binary classification scoring results}
\end{table}

These results demonstrate the model's effectiveness in detecting and ranking malicious behavior scores for users in a binary classification setting.

\section{Experimental result analysis}

\section{Conclusion}

In this paper, we have presented a general behavior pretraining model for anomaly detection and beyond. The model exhibits several advantages and has a wide range of practical applications. We summarize the key contributions and findings as follows:

\begin{enumerate}
    \item Model Advantages: The proposed model demonstrates strong scalability, making it suitable for scenarios with numerous features. It also exhibits few-shot learning capabilities, enabling it to effectively fit downstream tasks with limited samples. These attributes make the model a promising solution for imbalanced and cold-start problems.
    
    \item Broad Applicability: The model can be deployed in various scenarios, including vector-based retrieval for financial anti-fraud, fine-tuning on downstream tasks, and serving as features for supervised models in downstream tasks. This versatility allows the model to be applied across multiple domains, including risk control, marketing, and advertising.
\end{enumerate}

In conclusion, the proposed model is not only an anomaly detection model but also a general behavior pretraining model with the potential to transform a wide range of applications and domains.

\end{document}